\documentclass[letterpaper]{article} 
\usepackage{aaai24}  
\usepackage{times}  
\usepackage{helvet}  
\usepackage{courier}  
\usepackage[hyphens]{url}  
\usepackage{graphicx} 
\usepackage{natbib}  
\usepackage{caption} 
\usepackage{algorithm}
\usepackage{algorithmic}
\usepackage{times}
\usepackage{epsfig}
\usepackage{graphicx}
\usepackage{amsmath}
\usepackage{amssymb}
\usepackage{url}
\usepackage[utf8]{inputenc} 
\usepackage{url}            
\usepackage{booktabs}       
\usepackage{amsfonts}       
\usepackage{nicefrac}       
\usepackage{microtype}      
\usepackage{colortbl}
\usepackage{graphicx} 
\usepackage{amsmath}
\usepackage{listings}
\usepackage{multirow}
\usepackage{subfigure}
\usepackage{bbding}
\usepackage{sidecap}
\usepackage{threeparttable}
\usepackage{longtable}
\usepackage{mathtools}
\usepackage{enumitem}
\usepackage [english]{babel}
\usepackage [autostyle, english = american]{csquotes}
\usepackage{newfloat}
\usepackage{listings}
\DeclareCaptionStyle{ruled}{labelfont=normalfont,labelsep=colon,strut=off} 
\floatstyle{ruled}
\newfloat{listing}{tb}{lst}{}
\floatname{listing}{Listing}
\title{Decouple Content and Motion for Conditional Image-to-Video Generation}
\author{
    Cuifeng Shen\textsuperscript{\rm 1},
    Yulu Gan\textsuperscript{\rm 1},
    Chen Chen\textsuperscript{\rm 2},
    Xiongwei Zhu\textsuperscript{\rm 3},
    Lele Cheng\textsuperscript{\rm 3},
    Tingting Gao\textsuperscript{\rm 3}
    Jinzhi Wang\textsuperscript{\rm 1}
}
\affiliations{
    \textsuperscript{\rm 1}Peking University,
    \textsuperscript{\rm 2}Chinese Academy of Science,
    \textsuperscript{\rm 3}Kuaishou Technology \\
    \textit{cuifengc77@gmail.com}
}

\usepackage{bibentry}

\begin{document}
\maketitle

\begin{abstract}

The goal of conditional image-to-video (cI2V) generation is to create a believable new video by beginning with the condition, i.e., one image and text.The previous cI2V generation methods conventionally perform in RGB pixel space, with limitations in modeling motion consistency and visual continuity. Additionally, the efficiency of generating videos in pixel space is quite low.In this paper, we propose a novel approach to address these challenges by disentangling the target RGB pixels into two distinct components: spatial content and temporal motions. Specifically, we predict temporal motions which include motion vector and residual based on a 3D-UNet diffusion model. By explicitly modeling temporal motions and warping them to the starting image, we improve the temporal consistency of generated videos. This results in a reduction of spatial redundancy, emphasizing temporal details. Our proposed method achieves performance improvements by disentangling content and motion, all without introducing new structural complexities to the model. Extensive experiments on various datasets confirm our approach's superior performance over the majority of state-of-the-art methods in both effectiveness and efficiency.

\end{abstract}


\section{Introduction}

\begin{figure}[!t]
\centering 
\includegraphics[width=0.495\textwidth]{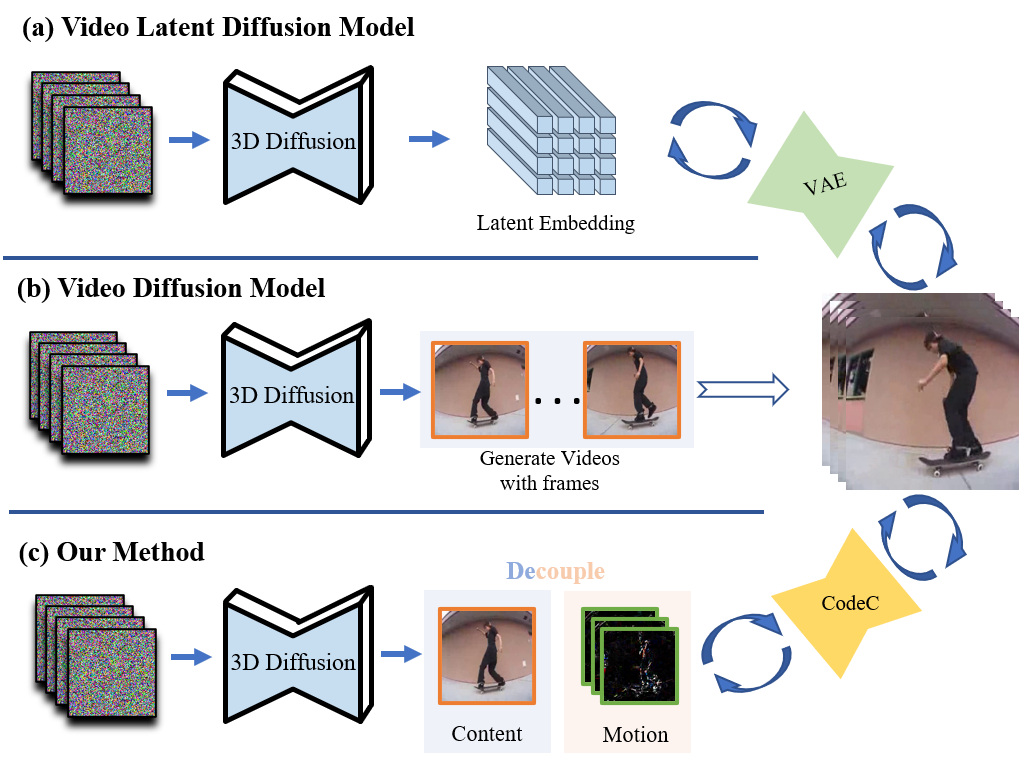} 
\caption{{\textbf{Motivations and our ideas}. Conventional methods (see in (b)), involve extending the RGB space with time sequences, resulting in limited memory efficiency and temporal coherence. Latent Diffusion Models employ a variational autoencoder for compression (depicted in (a)), enhancing efficiency but potentially reducing spatial quality and poor temporal coherence because temporal consistency hasn't been directly modeled. Our approach (refer to (c)) decouples the content and motion, capitalizing on existing temporal coherence in compressed video data, resulting in a memory-efficient and temporally consistent video generation approach.}}
\vspace{-0.3cm}
\label{Fig:teaser}
\end{figure}

Deep generative models have garnered wide attention, driven by the emergence of approaches like Generative Adversarial Models \cite{gan} and Diffusion Models \cite{ho2020denoising}. These models have recently demonstrated remarkable success in tasks such as image generation. In the field of the conditional Image-to-Video (cI2V), most works \cite{dalle, he2022lvdm, yu2022digan, vidm, mcvd, blattmann2023align} have showcased the potential and versatility of these cutting-edge techniques.

A line of works~\cite{yu2022digan, vidm, mcvd} in cI2V aims to directly extend images into videos by generating a series of image frames in RGB space (see in Fig \ref{Fig:teaser}(b)). However, they encounter the following challenges: (1) \textit{Information redundancy} of videos makes it difficult for a model to focus on the video's important temporal information. The slight variations present in each frame, combined with inherent redundancies within the video space, lead to a neglect of temporal details for the model when attempting to construct the video. This neglect hinders the ability of the model to focus on sequential video frames effectively. Consequently, the process of pixel-based generation can cause the model to disproportionately highlight the spatial content, thereby complicating the modeling of temporal motions. Achieving accurate and efficient generation of temporal information is notably demanding.
(2) \textit{Time consuming}. Generating the whole video for each frame in pixel space consumes significant resources, \textit{i.e., for a 16-frame 128x128x3 video, the target vector dimension would be 16x128x128x3.}

To address the high computational request of directly extending images into videos, some work \cite{blattmann2023align} encodes the image into latent space (see in Fig. \ref{Fig:teaser}(a)). However, this approach can cause a reduction in the quality of the frames due to the use of VAE, and still can not generate temporal consistent videos. Recent work LFDM \cite{ni2023conditional} uses a diffusion model to predict optical flow and then uses the optical flow in conjunction with the original image content to generate videos. However, optical flow cannot be easily inverted and is inaccurate, leading to the use of specialized flow predictors that may encounter local optima.

The approach of separating time and content information holds great potential. First, we can model the temporal information of videos individually, rather than considering all pixels together. Secondly, this approach allows us to save a considerable amount of computational cost. In comparison to LFDM, our method uses rigorous computation to model temporal information and is invertible.

Specifically, we first employ a simpler approach, named \textit{Decouple-Based Video Generation} (D-VDM) to directly predict the differences between two consecutive frames.
Then we propose the \textit{Efficient Decouple-Based Video Generation} (ED-VDM) method. We separate the content and temporal information of videos using a CodeC \cite{mpeg} to extract motion vectors and residual content. During training, we input the motion vectors, residual, and the first frame image together. The predictive model generates the motion vectors and residual output, and then we use the CodeC decoder to warp them with the image to restore the video. As we decouple the temporal and content information, during the prediction process, we aim for the joint probability distribution of input motion vectors and residual, while the model outputs the score of that distribution. Diffusion models have been proven effective in learning the score of joint distribution \cite{bao2022one}.

To recap, our main contributions are as follows:

\begin{itemize}[leftmargin=*,noitemsep,topsep=0pt]
\item Our proposed D-VDM decouples the video into content and temporal motions, enabling explicit modeling of the temporal motions. To model the decouplings separately, we use a diffusion-based method to model the temporal motions of a video and warp it to the given first frame. 

\item We investigate various compression techniques to decrease the spatial dimension of the temporal motion features. Our proposed ED-VDM, which employs an autoencoder to compress residuals, provides a 110x improvement in training and inference speed while maintaining state-of-the-art (SOTA) performance.

\item Extensive experiments are conducted, and the SOTA performance has been achieved on MHAD~\cite{mhad}, NATOPS~\cite{natops}, and BAIR~\cite{bair} datasets, demonstrating that our model can generate realistic and temporal consistent video. 
\end{itemize}

\section{Related work}

\subsection{Diffusion model}
Diffusion denoising probabilistic models (DDPMs)~\cite{sohl2015deep} learn to generate data samples
through a sequence of denoising autoencoders that estimate
the score~\cite{HyvrinenAapo2005EstimationON} of data distribution (a direction pointing toward higher density data).

Recently, diffusion probabilistic models~\cite{JostTobiasSpringenberg2015UnsupervisedAS, ho2020denoising, song2020score, karraselucidating} achieve remarkable progress in image generation \cite{rombach2022high, bao2022conditional}, text-to-image generation \cite{AlexNichol2023GLIDETP, AdityaRamesh2023HierarchicalTI, gu2022vector}, 3D scene generation \cite{BenPoole2023DREAMFUSIONTU}, image editing~\cite{ChenlinMeng2021SDEditIS, choi2021ilvr}. 

Our approach capitalizes on the exceptional capabilities of the diffusion model, but our key objective is to disentangle the video to enable the generation of motion features, resulting in improved temporal coherence. In addition, we strive to minimize spatial redundancy during generation tasks that involve temporal difference features.

\subsection{Video Generation with diffusion model}
Video generation aims to generate images with time sequences. 
In the context of diffusion-based model design, VDM~\cite{he2022lvdm} extended a 2D U-net architecture with temporal attention. Make-a-Video~\cite{makeavideo}, ImageN-Video \cite{Ho2022ImagenVH}, and Phenaki~\cite{phenaki} have applied video diffusion models to the generation of high-resolution and long-duration videos, leveraging high computational resources. To get rid of high GPU memory consumption, MCVD~\cite{mcvd} generates videos in a temporal autoregressive manner to reduce architecture redundancy. PVDM~\cite{pvdm} and LVDM~\cite{he2022lvdm} propose a 3D latent-diffusion model that utilizes a VAE to compress spatiotemporal RGB pixels within the latent space. However, different from motion difference features, spatial pixels usually only contribute a limited 8$\times$ spatial compression ratio.

Unlike previous methods that generate videos in RGB space, our proposed approach transfers the target space into a compressed space, resulting in significant performance improvements and a notable spatial downsample ratio (16$ \times$).

\subsection{Video Compression \label{sec:vd}}
Video compression is to reduce the amount of data required to store or transmit a video by removing redundant information. 
MPEG-4~\cite{mpeg} is one of the most commonly used methods for video compression. MPEG-4 utilizes motion compensation, transform coding, and entropy coding to compress the video. 
CodeC categorizes video into I-frames, P-frames, and optional B-frames. I-frames are standalone frames with all image info. P-frames encode frame differences via motion vectors and residuals for object movement and image variance. Note that Traditional CodeCs compress P-frames with DCT, quantization, and entropy, an unsuitable format. To address this, we apply a fixed-length VAE for motion compression.

\subsection{Computer Vision on Decoupled Videos}
Video decoupling separates spatial and temporal information in videos to improve video understanding and compression. Recent works, such as~\cite{cheng2021modular, yang2022decoupling, huang2021self}, have highlighted the importance of video decoupling in video perception and understanding. Moreover, some methods have achieved impressive performance in video understanding by decoupling the video into spatial information (the first frame) and sequential information (motion vectors and residuals). \cite{wu2018compressed} directly train a deep network on compressed video, which simplifies training and utilizes the motion information already present in compressed video. \cite{huang2021self} use key-frames and motion vectors in compressed videos as supervision for context and motion to improve the quality of video representations. Different to previous methods using decoupled features as conditioning, our approach directly generates those features.

\section{Decouple Content and Motion Generation}
The goal of conditional image-to-video is to generate a video given the first frame and condition. Assume $\mathbf{s}\thicksim \mathcal{N}(0, I)$ is a Gaussian noise with the shape of  $N \times H \times W \times C$ where $N $, $ H$, $W $, and $ C$ represents the number of frames, height, width, and channel respectively. Denote $x_0$ as the first frame of a video clip, and $\mathbf{x_0}=\{x_0, x_1,..., x_K\} $ represents the video clip which has the same shape as the Gaussian noise. During training, the diffusion model learns the score of the video distribution. During sampling, starting with the initial frame $x_0$ and condition $y$ we generate a video clip $\mathbf{\hat{x}_0}=\{\hat{x}_0, \hat{x}_1,..., \hat{x}_K\} $ from the learned distribution, beginning with Gaussian noise $\mathbf{s}$. Based on the datasets, we only use text labels as the condition $y$. In this section, we first introduce the preliminary diffusion models and explain our proposed D-VDM and ED-VDM methods in detail. 

\subsection{Diffusion model}
\begin{figure*}[h] 
\centering 
\includegraphics[width=.97\textwidth]{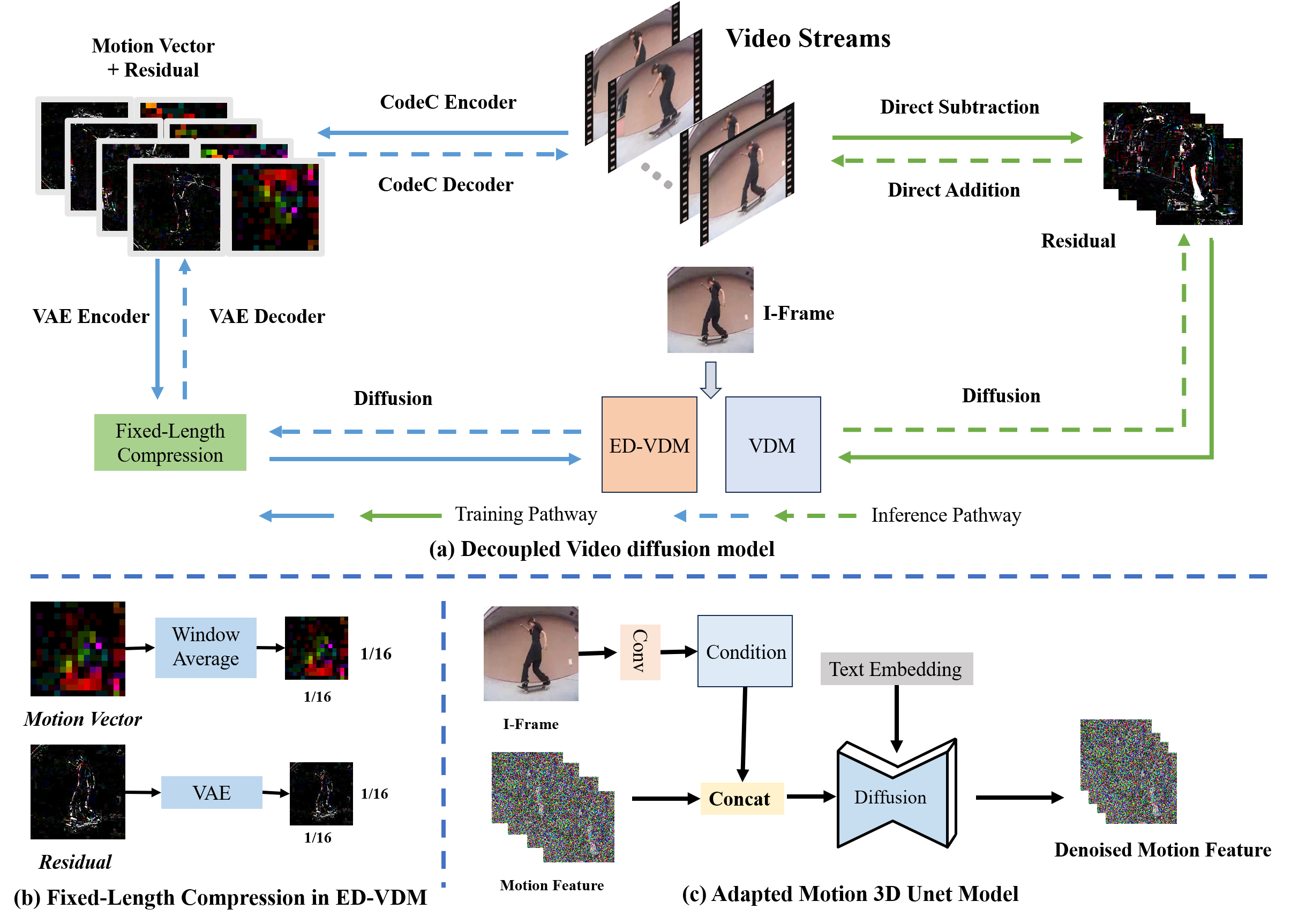} 
\caption{{ \textbf{Illustration of our proposed decoupled video diffusion model.} (a) \textbf{Pipeline.} The green pathway represents the Decoupled Video Diffusion Model (D-VDM), which directly generates motion features in the compressed video domain, while the blue pathway illustrates the Efficient Decoupled Video Diffusion Model (ED-VDM), which includes a reversible compression function. (b) \textbf{Compression techniques used in the ED-VDM model.} Since the separated motion vectors and residuals are of unequal lengths, it is necessary for us to apply equal-length processing to both components. (c) \textbf{The architecture of the 3D U-Net.} We employed the 3D U-Net architecture in both models. }}
\vspace{-0.1cm}
\label{Fig.main}
\end{figure*}

Denoising Diffusion Possibility Model (DDPM)~\cite{ho2020denoising} consists of a forward process that perturbs the data to a standard Gaussian distribution and a reverse process that starts with the given Gaussian distribution and uses a denoising network to gradually restore the undisturbed data structure. 

Specifically, consider $\mathbf{x}_0$ as a data sample from the distribution $\mathbf{x}_0\thicksim q(\mathbf{x_0})$, representing a video clip in this study. We denote $T$ as the total step of the perturbation. 
In the forward process, DDPM produces a Markov chain $\mathbf{x}_0,..., \mathbf{x}_T$ by injecting Gaussian noise $\mathcal{N}(0, I)$ to $\mathbf{x}_0$, that is:

\begin{equation}
    q(\mathbf{x}_t|\mathbf{x}_{t-1}):=\mathcal{N}(\mathbf{x_{t-1}};\sqrt{\alpha_t}\mathbf{x}_{t-1}, \beta_tI).
    \label{eq.3.1.1}
\end{equation}

\begin{equation}
    q(\mathbf{x}_{1:T}|\mathbf{x}_0)=\prod \limits_{t=1}^T q(\mathbf{x}_t|\mathbf{x}_{t - 1}),
    \label{eq.3.1.2}
\end{equation}

\noindent where $\alpha_t=1-\beta_t$ and $\beta_t$ is the noise schedule. 

Regarding the denoising process, when the value of $T$ is sufficiently large, the posterior distribution $q\left(\mathbf{x}_{t-1} \mid \mathbf{x}_t\right)$ can be approximated as a Gaussian distribution. The reverse conditional probability can be computed using Bayes' rule conditioned on $x_0$:
\begin{equation}
q\left(\mathbf{x}_{t-1} \mid \mathbf{x}_{t}, \mathbf{x}_0 \right):=\mathcal{N}\left(x_t ; \mathbf{\hat{\mu}}(\mathbf{x}_{t}, \mathbf{x}_0), \hat{\sigma}\right),
\label{eq:backward1}
\end{equation}
where $\mathbf{\hat{\mu}}(\mathbf{x}_{t}, \mathbf{x}_0)$ is obtained as:
\begin{equation}
\mathbf{\hat{\mu}}(\mathbf{x}_{t}, \mathbf{x}_0) = \cfrac{\sqrt{\alpha_t}(1-\bar{\alpha}_{t-1})}{1-\bar{\alpha}_{t}} \mathbf{x}_t + \cfrac{\sqrt{\bar{\alpha}_{t-1}} \beta_t}{1 - \bar{\alpha}_t} \mathbf{x}_0,
\label{eq:backward2}
\end{equation}
Moreover, by integrating equation ~\ref{eq.3.1.2}, predicting the original video $\mathbf{x}_0$ is equivalent to predict the noise $\epsilon$ added in $\mathbf{x}_t$:
\begin{equation}
\mathbf{\hat{\mu}}(\mathbf{x}_{t}, \mathbf{x}_0) = \frac{1}{\sqrt{\alpha_t}} \Big( \mathbf{x}_t - \frac{1 - \alpha_t}{\sqrt{1 - \bar{\alpha}_t}} \boldsymbol{\epsilon}_t \Big),
\label{eq:backward3}
\end{equation}

Hence, to estimate $\mathbf{\hat{\mu}}(\mathbf{x}_{t}, \mathbf{x}_0)$, we need to learn the function $\mu_{\theta}(\mathbf{x}_{t}, t)$. We can achieve this by directly learning the noise $\epsilon_{\theta}(\mathbf{x}_t, t)$:
\begin{equation}
    \mathbb{E}_{t\thicksim \mathcal{U}(0,T),\mathbf{x}_0\thicksim q(\mathbf{x}_0),\epsilon\thicksim \mathcal{N}(0,1)}[\lambda(t)\lVert\epsilon-\epsilon_\theta(\mathbf{x}_t, t)\rVert^2],
\end{equation}

\subsection{Decoupled Video Diffusion Model}\label{Decouple-based vg}
Video diffusion models aim to use DDPM to estimate the score of the video distribution $\mathbf{v}_0 \sim q(\mathbf{v}_0)$, where $\mathbf{v}_0 = \{v_0, v_1,..., v_K\}$ belongs to the RGB pixel space $\mathbb{Z}^{3\times K \times W \times H}_{[0,255]}$, and $K$ is the frame number, $W$ and $H$ is frame width and height respectively. The denoising 3D Unet is designed to learn a denoising parameter $\epsilon_\theta(\mathbf{v}_t, t)$.

\begin{figure}[tb] 
\centering 
\includegraphics[width=0.47\textwidth]{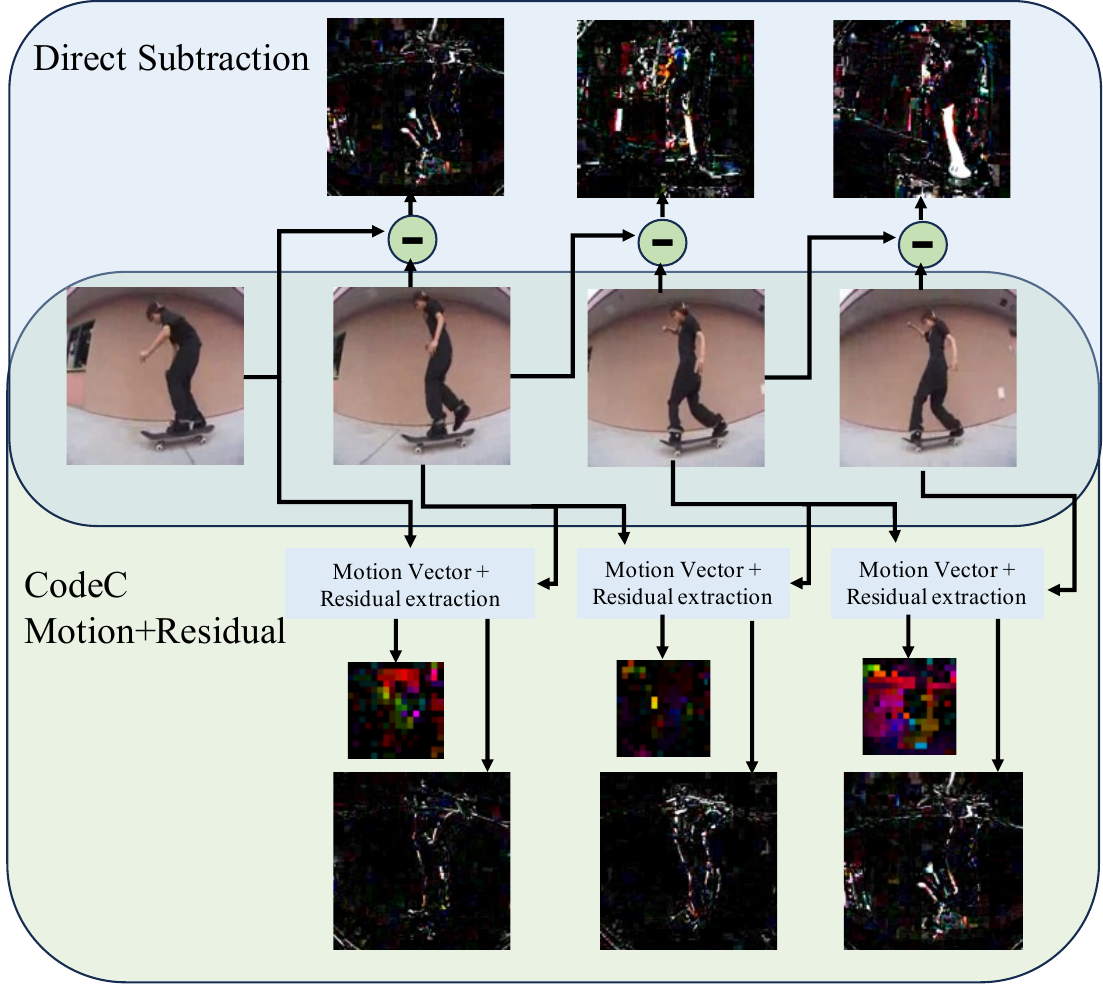}
\caption{{Different ways to represent video temporal feature between frames.  \textbf{Frame difference used by D-VDM}, a simple technique, calculates direct frame discrepancies, encompassing both fundamental and advanced temporal alterations. D-VDM findings reveal its potency in refining the temporal consistency of a generated video.  \textbf{Motion Vector and Residual used by ED-VDM}, as in H.264, disentangles temporal shifts into intermediate motion blocks and pixel residuals. Notably, correlated with motion vectors, these residuals offer a sparse representation with potent compression potential. In our experiments, ED-VDM attains an impressive 110x compression ratio.}
\vspace{-0.2cm}
}

\label{Fig.ddiffvis}
\end{figure}

One simple approach to decouple a video into spatial and temporal representations, as illustrated in Figure \ref{Fig.ddiffvis}, is to retain the first frame and then compute the differences between it and the subsequent frames, noted as $\hat{\mathbf{v}}_0 \in \mathcal{V}(\hat{\mathbf{v}}_0)$, where $\hat{\mathbf{v}}^n_0 = \mathbf{v}^n_0 - \mathbf{v}^{n-1}_0, \hat{v}^{2...n}_0\sim \mathbb{Z}^{3\times K-1 \times W \times H}_{[-255,255]}$.

To align the difference with the first frame and provide spatial information, low-level semantic information of the first frame is incorporated into the learning target. This is achieved by using a ResNet~\cite{he2016deep} bottleneck module to encode the first frame (denoted as $\tau_\theta(v_0)$) and concatenating it to the learning target along the channel dimension. Subsequently, the learning objective is slightly modified as
\begin{equation}
    L=\mathbb{E}_{t,\hat{v}_0\thicksim \mathcal{V}(\hat{v}_0),\epsilon\thicksim \mathcal{N}(0,1)}[\lambda(t)\lVert\epsilon-\epsilon_\theta(\hat{v}_t, t, \tau_\theta(v_0))\rVert^2],
\end{equation}
where $\tau_\theta$ and $\epsilon_\theta$ is jointly optimized. 

\subsection{Efficient Decoupled Video Diffusion Model \label{Efficient Decouple-based vg}}

As discussed in the previous section, another efficient representation of decoupled video can be achieved through I-frames and P-frames. P-frames comprise motion vectors and residuals, as depicted in Figure \ref{Fig.ddiffvis}.

We follow the H.264 protocol to obtain the motion vector $\mathbf{m}$ and residuals $\mathbf{r}$ from a video tube $\mathbf{v}$, using a reversible function $f(\mathbf{v}) = <\mathbf{m}, \mathbf{r}>$:

Let $\mathbf{v}^{n}$ and $\mathbf{v}^{n-1}$ be the current and the previous frame, respectively. We divide $\mathbf{v}^{n-1}$ into non-overlapping macroblocks of size $16 \times 16$ pixels, denoted as ${B_{i}}$, where $i$ represents the index of the macroblock. To obtain the motion vector $\mathbf{m}_{i}$ for each macroblock $B_{i}$, we search for a corresponding block $B'_{i}$ in the current frame $\mathbf{v}^{n}$ that is similar to $B_{i}$. This can be achieved by minimizing the sum of absolute differences between the two blocks, which can be formulated as:
$$\mathbf{m}_i=\underset{\mathbf{u}, \mathbf{w}}{\operatorname{argmin}} \sum_{j, k}\left|B_{i}(j, k) -B'_{i}(j+\mathbf{u}, k+\mathbf{w})\right|$$
Once the motion vector $\mathbf{m}_{i}$ is obtained, the residual $\mathbf{r}$ can be calculated as the difference between the previous macroblock $B_{i}$ and the motion-compensated block $B'_{i}$ in the current frame, i.e.,
$\mathbf{r}_{i}=B_{i}-B'_{i}(\mathbf{m}_{i})$, The motion vectors and residuals for all macroblocks can then be combined to form the P-frame.

Motion vectors that are used to represent the motion information of $16\times16$ blocks in the video frames contain identical numbers, and thus can achieve a high spatial compression ratio of $256\times$. Residuals, on the other hand, have the same spatial size as the video frames, but contain less information than the original frames and thus can be compressed efficiently. For the residual compression, we utilize a Latent Diffusion~\cite{rombach2022high} autoencoder to compress the residuals into a latent space. In order to match the spacial dimension of the motion vector, the residual downsampling rate is set equal to the motion vector compression rate and in our case is $16\times$.
Specifically, given a residual $\mathbf{r} \in \mathbf{z}^{3 \times W \times H}_{[-255,255]}$, the encoder $\mathcal{E}$ encodes the residual to a latent space $z=\mathcal{E}(\mathbf{z}) \in \mathbb{R}^{16 \times \frac{W}{16} \times \frac{H}{16}} $, and the decoder $\mathcal{D}$ could reconstruct the image from the latent $\mathbf{r}'=\mathcal{D}(\mathcal{E}(\mathbf{r}))$. We use $L_1$ as our objective to train the autoencoder. 

The dense representation of motion vector and residual with the same spatial resolution enables us to concatenate them channel-wise as $[\mathbf{m}, \mathbf{r}]$. 
We can uniformly sample the time steps $t$ to learn the joint distribution of $[\mathbf{m}, \mathbf{r}]$ with the following objective function:
\begin{equation}
    \begin{aligned}
    L &=\mathbb{E}_{t, \mathbf{m},\mathbf{r} =f(v_0), v_0\thicksim \mathcal{V},\epsilon\thicksim \mathcal{N}(0,1)}\left[\lambda(t) \operatorname{mse} \right] \\
    \operatorname{mse} &= 
    \lVert\epsilon - \epsilon_\theta(\mathbf{m}_t, \mathcal{E}(\mathbf{r}_t), t, \tau_\theta(v_0^0))\rVert^2 . 
    \end{aligned}
\end{equation}

\begin{table*}[t]
\begin{minipage}[t]{1\linewidth}
\footnotesize
\centering
\renewcommand\arraystretch{1.2}
\setlength\tabcolsep{16pt}
\makeatletter\def\@captype{table}
\caption{\textbf{Quantitative comparison of conditional Image-to-Video generation on MHAD and NATOPS datasets.} We compare FVD, sFVD, and cFVD on 16 frames clip. The 64 and 128 in the subscript indicate that the resolution of synthesized video frames is $64 \times 64$ and $128 \times 128$, respectively.}

\begin{tabular}{l|ccc|ccc}
        \hline
        Method  & 
       \multicolumn{3}{c}{\textit{MHAD}}  &  \multicolumn{3}{c}{\textit{NATOPS}} \\ \cline{2-7}
        &FVD$\downarrow$ & cFVD$\downarrow$ & sFVD$\downarrow$ &FVD$\downarrow$ & cFVD$\downarrow$ & sFVD$\downarrow$  \\ \hline 
        ImaGINator (WACV 2020)& 889.48  & 1406.56  & 1175.74&721.17&1122.13 &1042.69 \\
        VDM (Arxiv 2022)& 295.55  &  531.20 & 398.09&169.61&410.71&350.59 \\
        $\text{LDM}_{64}$(CVPR 2022) & 280.26  & 515.29 & 427.03&251.72&506.40& 491.37\\
        $\text{LFDM}_{64}$(CVPR 2023) &152.48 &339.63 &242.61 &160.84&376.14&324.45\\        
        $\textbf{D-VDM}_{64} $ &  \textbf{145.41} &  \textbf{308.33} & \textbf{244.73}
        & \textbf{152.19}& \textbf{358.47}& \textbf{266.53} \\ \hline 
        $\text{LDM}_{128}$ (CVPR 2022) &337.43  & 594.34 & 497.50 &344.81&627.84&623.13 \\
        $\text{LFDM}_{128}$ (CVPR 2023) &214.39&426.10 &\textbf{328.76} &195.17&423.42 &369.93 \\ \hline
        
         $\textbf{ED-VDM}_{128} $ (110$\times$ speedup)  &  \textbf{204.17}  &  \textbf{389.70} & 348.10 & \textbf{179.65}& \textbf{373.23}& \textbf{351.26} \\\hline 

\end{tabular}
\label{table:highres}
\end{minipage}
\vspace{-0.1cm}
\end{table*}

\section{Experiments}

\subsection{Datasets and Metrics}

\subsubsection{Datasets}
We conduct our experiment on well-known video datasets used for image-to-video generation: MHAD~\cite{mhad}, NATOPS~\cite{natops}, and BAIR~\cite{bair}. 
The MHAD human action dataset comprises 861 video recordings featuring 8 participants performing 27 different activities. This dataset encompasses a variety of human actions including sports-related actions like bowling, hand gestures such as 'draw x', daily activities like transitioning from standing to sitting, and workout exercises like lunges. For training and testing purposes, we've randomly picked 602 videos from all subjects for the training set and 259 videos for the testing set.
The NATOPS aircraft handling signal dataset encompasses 9,600 video recordings that involve 20 participants executing 24 distinct body-and-hand gestures employed for interaction with U.S. Navy pilots. This dataset features common handling signals like 'spread wings' and 'stop'. We have arbitrarily chosen 6720 videos from all subjects for the training phase, while the remaining videos are for the testing phase. Detailed data preprocess methods are listed in Appendix.  


\subsubsection{Metrics}
For evaluation metrics of the text conditional image to video task, we follow the protocol proposed by LFDM~\cite{ni2023conditional} and report the Fr\'echet Video Distance (FVD) ~\cite{fvd}, class conditional FVD (cFVD) and subject conditional FVD (sFVD) for MHAD and NATOPS datasets. 
FVD utilizes a pre-trained I3D~\cite{carreira2017quo} video classification network from the Kinetics-400~\cite{kay2017kinetics} dataset to derive feature representations of both real and generated videos. Following this, the Frechet distance is computed to measure the difference between the distributions of the real and synthesized video features.
The cFVD and sFVD evaluate the disparity between the actual and generated video feature distributions when conditioned on the same class label $y$ or the identical subject image $x_0$, respectively.
In addition, for the image-to-video task, we report the FVD score on BAIR datasets. 
All evaluation is conducted on 2048 randomly selected real and generated 16 frames video clips following the protocol proposed by StyleGAN-V~\cite{StyleGAN}.

\subsection{Implementation Details}

we use a conditional 3D U-Net architecture as the denoising network, . We directly apply the multi-head self-attention~\cite{cheng2016long} mechanism to the 3D video signal. The embedding $e$ of the condition $y$ is concatenated with the time step embedding.
Additionally, we use a ResNet\cite{he2016deep} block to encode the first frame as a conditional feature map and provided it to $\epsilon_\theta$ by the concatenation with the noise $\epsilon$. In ED-VDM, The feature map of the first frame is downsampled $16\times$ to match the size of the motion feature, and in D-VDM the feature map remains the original size. We use the pre-trained CLIP~\cite{clip} to encode text $y$ as text embedding, and we adopt the classifier-free guidance method in the training process. Detailed U-net structures of ED-VDM and D-VDM can be found in the supplementary material.  

For ED-VDM, we compress the motion vector according to the block size (values in the same block inside the motion vector are the same), and we employ a VAE~\cite{rombach2022high} with slight KL-regularization of $ 1e^{-6}$ to encode the residual into a $16\times16\times16$ latent representation. Detailed model architectures are listed in the supplementary material. For different datasets, we take the middle 4 residuals out of 16 in a video clip as the training set to train our VAE model. 

\subsection{Baselines}
We compare our approach with the most recent diffusion-based methods including MCVD~\cite{mcvd}, CCVD~\cite{lemoing2021ccvs}, and VDM~\cite{ho2022video} on image-to-video tasks in BAIR datasets. 
We also compare our method with LFDM\cite{ni2023conditional}, VDM~\cite{ho2022video}, and LDM~\cite{he2022latent} on MHAD and NATOPS for text condition image-to-video tasks.  We collect the performance scores of the above methods from their original paper and LFDM paper. 

\begin{figure*}[t] 
\centering 
\captionsetup{width=.99\textwidth}
\includegraphics[width=0.99\textwidth]{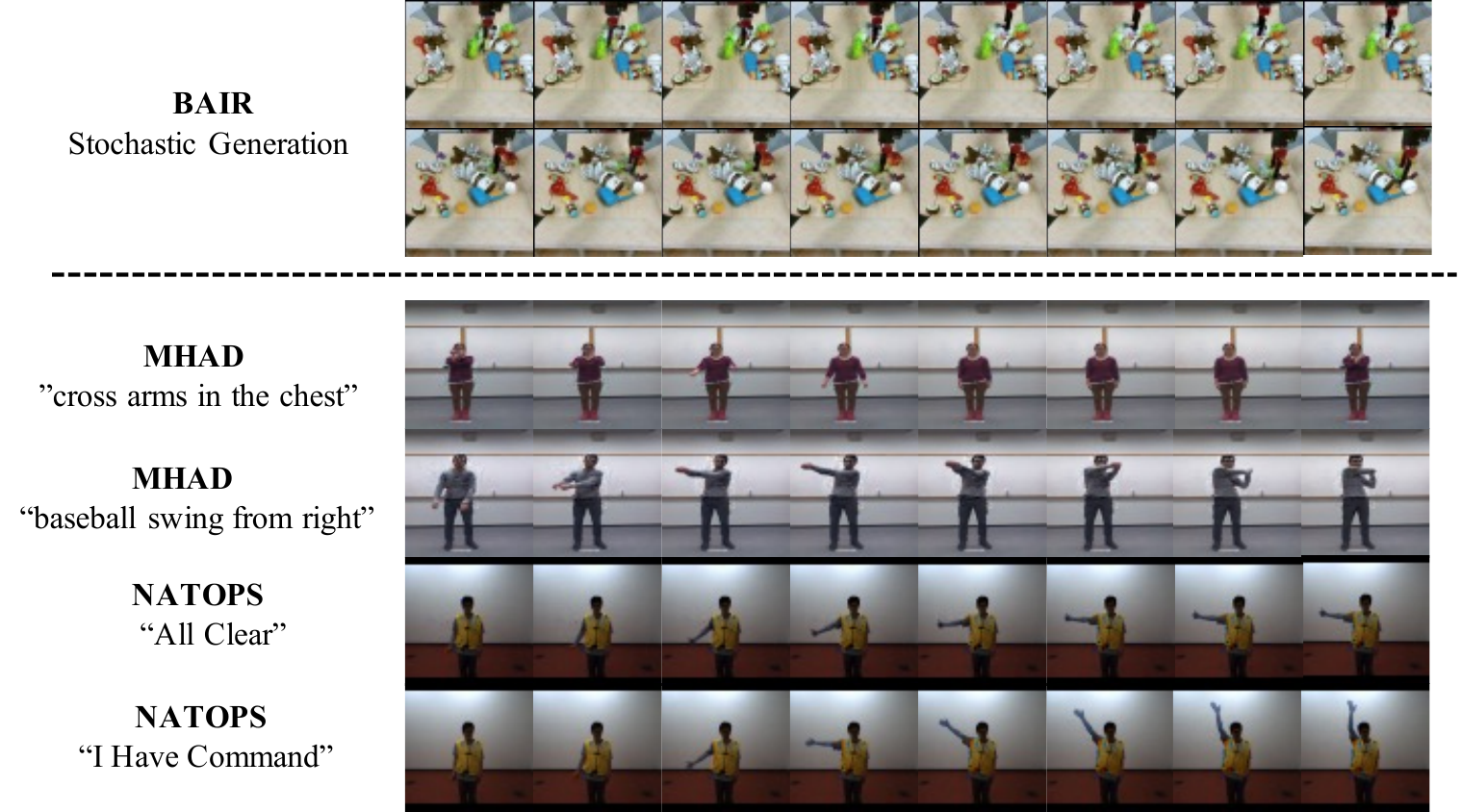} 
\caption{{ \textbf{Selected samples on BAIR, NATOPS, and MHAD dataset.} First two rows are the results of unconditionally generation results on BAIR, and the down four rows are text conditional generation results on MHAD and NATOPS. The visualization results show our method generated realistic and temporally consistent video frames.}}
\vspace{-0.2cm}
\label{Fig.tachi32}
\end{figure*}

\subsection{Main Results}

\textbf{Stochastic Image to Video Generation.} Table \ref{table:bair} shows the quantitative comparison between our method and baseline methods for the image-to-video (I2V) task on the BAIR dataset with $64\times64$ resolution.
By simply changing the target space from RGB pixels to image residuals, D-VDM improved the previous best FVD from 66.9 to 65.5. By utilizing motion vector and residual, ED-VDM achieves a high compression rate and a competitive performance. 
For image quality, our D-VDM method surpasses previous methods on PSNR, SSIM, and LPIPS from 16.9 to 17.6, 0.780 to 0.799, and 0.122 respectively. The image quality of the generated video by ED-VDM is slightly lower than the SOTA method but with a much higher speed advantage. We assume that due to the low image resolution, the compressed temporal latent only has a spatial size of $4\times4$ which can not contain enough information to restore the original video. Further experiments on high-resolution videos demonstrate the superiority of our ED-VDM method.


\begin{table}[t]
    \centering
    \footnotesize
    \renewcommand\arraystretch{1.2}
    \setlength\tabcolsep{14pt}
    \caption{\textbf{The upper bound of our ED-VDM method.} R-FVD score is evaluated with 2,048 samples. PSNR and SSIM are evaluated on an average of 16 frames with 100 samples.}
    \begin{tabular}{cccccc}
        \toprule
        ~ & R-FVD$\downarrow$ &PSNR$\uparrow$ & SSIM$\uparrow$\\ 
        \midrule
        128-MHAD& 130.62&31.80  & 0.95\\
        128-NATOPS& 131.91&31.60  &0.96 \\
         \bottomrule
    \end{tabular}
    \label{table:recon_quant}
    \vspace{-0.2cm}
\end{table}

\textbf{Conditional Image to Video Generation.}
We conduct experience on NATOPS and MHAD following the protocols proposed by LFDM~\cite{ni2023conditional}. Our D-VDM achieved remarkable results on MHAD and NATOPS at $64\times64$ resolutions, outperforming all previous SOTA methods. In specific, D-VDM achieves an FVD score of \textbf{145.41} on MHAD and \textbf{152.19} on NATOPS. Considering the effect of the text condition and the subject image, we further report the cFVD and sFVD scores of our method, and our method achieves SOTA performance. The results provide further evidence of the effectiveness of our proposed decoupled-based method. These results validate our motivation and demonstrate that merely changing the target space can greatly improve model performance.
We conducted further experiments with the ED-VDM method on NATOPS and MHAD with larger resolutions. Our results are presented in Table~\ref{table:highres}, and ED-VDM achieved comparable results at 128 resolutions with 110 times speedup than VDM~\cite{he2022lvdm}. Specifically, our proposed ED-VDM achieved an FVD score of \textbf{204.17} on MHAD and \textbf{179.65} on NATOPS which surpasses all previous methods. For the different text conditions and subject images, our method achieves SOTA performance on both sFVD and cFVD.

For qualitative results, Figure \ref{Fig.tachi32} illustrates the video generation samples from our D-VDM and ED-VDM methods on BAIR, MHAD, and NATOPS datasets. The figure demonstrates that our proposed approach can generate realistic and temporally consistent videos on three datasets. With the text condition on dataset MHAD and NATOPS, generated videos achieve a strong correlation with the text condition. Furthermore, using ED-VDM, we can still generate high-fidelity videos with comparable quality, which leverage a 110 times training and inference efficiency.

\section{Analysis} 

\subsection{Reconstruction Quality}
\begin{figure}[t]
    \centering 
    \includegraphics[width=.47\textwidth]{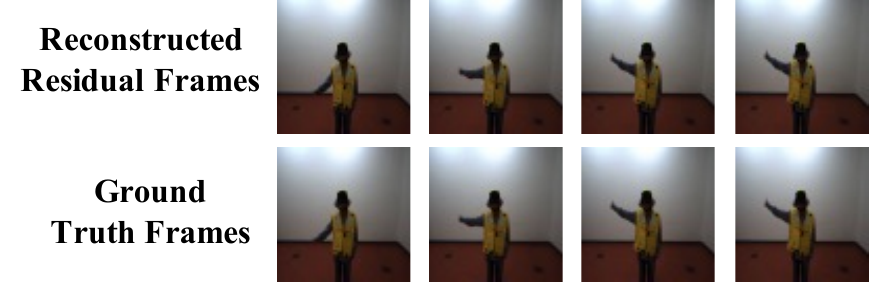} 
    \caption{ Video quality on residual reconstruction.}
    \label{Fig.reconstruct}
    \vspace{-0.2cm}
\end{figure}

Table \ref{table:recon_quant} summarizes the results of the reconstruction quality of our residual autoencoder.  We use the R-FVD, which indicates FVD between reconstructions and the ground-truth real videos, peak-signal-to-noise ratio (PSNR), and structural similarity index measurement (SSIM) to evaluate the image quality with residual reconstruction. All evaluations are conducted on randomly selected reconstructed videos and real videos. Quantitative results in Figure \ref{Fig.reconstruct} show that our residual reconstruction method achieves a small image quality degradation.

\begin{table}[t]
\footnotesize
\centering
\renewcommand\arraystretch{1}
\setlength\tabcolsep{4pt}
\makeatletter\def\@captype{table}
\caption{\textbf{Image-to-Video Generation
 Results on BAIR dataset.} Our method surpasses the SOTA methods with regard to FVD score.}

\begin{tabular}{llcccc}
        \toprule
        BAIR (64x64) & FVD$ \downarrow$ & PSNR $\uparrow$ & SSIM $\uparrow$ & LPIPS $\downarrow$ \\ 
        \midrule
        CCVS & 99.0&-&0.729&- \\
        MCVD& 89.5 &16.9&0.780&-\\
        VDM&66.9&-&-&-\\
         \textbf{D-VDM}  &  \textbf{65.5}& \textbf{17.6}& \textbf{0.799}& \textbf{0.122} \\
         \textbf{ED-VDM} (110×speedup)  &  92.4& 16.0& 0.775& 0.132\\
         \bottomrule
\end{tabular}

\label{table:bair}
\end{table}

\begin{table}[h]%
    \centering
    \footnotesize
    \renewcommand\arraystretch{1.2}
    \setlength\tabcolsep{8pt}
    \caption{FLOPs and memory usage for our model to train on 1 batch of $16\times128\times128\times3 $ resolution videos.}
    \vspace{-0.3cm}
    
    \begin{tabular}{cccccc}
        \toprule
        Method & FLOPs($\times10^{9}$)&Memory (GB) \\ 
        \midrule
        VDM \cite{ho2022video} &8814 &11.56 \\
        LFDM \cite{ni2023conditional} & 627 & 6.69   \\
        \midrule
        \textbf{D-VDM}&8611 &11.49 \\
         \textbf{ED-VDM} & \textbf{78} & \textbf{3.47}  \\
         \bottomrule
    \end{tabular}
    \label{table:speed}
\end{table}

\subsection{Speed Comparison}
As shown in Table \ref{table:speed}, we evaluate the FLOPs and memory consumption to train on $128\times128\times3$ video clips of different methods.  
Since D-VDM directly uses a 3D U-net to train on original video frames, it has approximately the same FLOPs and memory as the video diffusion model (VDM). With the residual compression and motion vector, our ED-VDM model achieves a $256\times$ spatial compression rate than VDM and $\sim110\times$, $\sim8\times$ better computation efficiency than VDM and LFDM, respectively.

\subsection{Compression Method Exploration}
To compress the residual to match the dimension with the motion vector, we evaluated two methods to compress and reconstruct, including Discrete Cosine Transformation (DCT) and autoencoder compression. Table \ref{table:recon} shows the image reconstruction quality of different approaches. We can see that the adopted autoencoder achieves better reconstruction quality in both metrics.

\begin{table}[H]
    \footnotesize
    \centering
    \renewcommand\arraystretch{1}
    \setlength\tabcolsep{20pt}
    \renewcommand\arraystretch{1.2}
    \setlength\tabcolsep{16pt}
    \caption{Image quality comparison between our proposed autoencoder method and traditional DCT method.}
    \vspace{0.1cm}
    \label{table:recon}
    \begin{tabular}{cccccc}
        \toprule
        128-MHAD-101 & PSNR$\uparrow$ & SSIM$\uparrow$\\\midrule
        Autoencoder&31.80&0.96\\
        DCT&17.08&0.85\\ \bottomrule
    \end{tabular}
\end{table}

\section{Conclusion}
This paper demonstrates that transforming the target space from RGB pixels to the spatial and temporal features used in video compression can significantly improve the temporal consistency and computational efficiency of video generation. We propose Decoupled Video Diffusion Model (D-VDM), which achieves SOTA performance on various video generation tasks by decoupling the video into key frame and temporal motion residuals. Furthermore, our proposed ED-VDM further takes advantage of the sparsity in the motion compensation features to achieve comparable SOTA results with notable speedup (110$\times$). These results demonstrate the effectiveness of our decouple-based approach and open up possibilities for future work in video generation research.

\bibliography{aaai24}

\end{document}